\pdfoutput=1

\documentclass[11pt]{article}

\usepackage{emnlp2021}

\usepackage{times}
\usepackage{latexsym}

\usepackage[T1]{fontenc}

\usepackage[utf8]{inputenc}

\usepackage{microtype}

\usepackage{multirow}
\usepackage{graphicx}
\usepackage{enumitem}
\usepackage{booktabs}
\usepackage{color}
\usepackage{subfig}
\usepackage{gb4e}
\noautomath

\title{Patterns of Lexical Ambiguity in Contextualised Language Models}

\author{{Janosch Haber \and Massimo Poesio}\\
  Queen Mary University of London and The Alan Turing Institute\\
  \texttt{\{j.haber|m.poesio\}@qmul.ac.uk}\\
}

\begin{document}
\maketitle
\begin{abstract}

One of the central aspects of contextualised language models is that they should be able to distinguish the meaning of lexically ambiguous words by their contexts. In this paper we investigate the extent to which the contextualised embeddings of word forms that display multiplicity of sense reflect traditional distinctions of polysemy and homonymy.  To this end, we introduce an extended, human-annotated dataset of graded word sense similarity  and co-predication acceptability, and evaluate how well the similarity of embeddings predicts similarity in meaning.

Both types of human judgements indicate that the similarity of polysemic interpretations falls in a continuum between identity of meaning and homonymy. However, we also observe significant differences within the similarity ratings of polysemes, forming consistent patterns for different types of polysemic sense alternation. Our dataset thus appears to capture a substantial part of the complexity of lexical ambiguity, and can provide a realistic test bed for contextualised embeddings.

Among the tested models, BERT Large shows the strongest correlation with the collected word sense similarity ratings, but struggles to consistently replicate the observed similarity patterns. When clustering ambiguous word forms based on their embeddings, the model displays high confidence in discerning homonyms and some types of polysemic alternations, but consistently fails for others. 




\end{abstract}

\section{Introduction}


Capturing lexical ambiguity
has  been a driving factor in the development of  contextualised language models \cite[e.g.][]{Peters2018DeepRepresentations, devlin2018bert}.
Evaluating their performance, much of the focus 
has been on homonymy, a variety of 
multiplicity of meaning 
exemplified by word forms such \textit{match} in (\ref{ex:homonym}),  whose  different meanings are entirely unrelated.
\begin{exe}
\ex
a. The match burned my fingers. \\
b. The match ended without a winner.
\label{ex:homonym}
\end{exe}
And indeed, research such as \citep{wiedemann2019does, loureiro-jorge-2019-language, blevins-zettlemoyer-2020-moving} has achieved  promising results on using contextualised language models to disambiguate homonyms.
But homonymy is not the only form 
lexical ambiguity 
can take 
\citep{Pinkal:1995, Cruse2009PolysemyViewpoint,poesio_forthcoming}:
in polysemy, word forms 
like \textit{school} in (\ref{ex:polyseme})
can elicit different distinct but related senses 
\cite{lyons_1977}. 
\begin{exe}
\ex
a. They agreed to meet at the school. \\ 
b. The school has prohibited drones. \\
c. The school called Tom's parents. 
\label{ex:polyseme}
\end{exe}
Polysemy is in fact much more common than homonymy, and most words can be considered polysemous to some degree \cite{Rodd_2004, Falkum2015PolysemyCP,poesio_forthcoming} --however,
the ability of contextualised language models to capture this phenomenon has been studied much less.  
In this paper, we 
shift the focus to polysemy proper, and investigate how well contextualised language models capture graded word sense similarity as observed in human annotations. 


It is important to carefully distinguish polysemy from homonymy, as 
multiplicity of meaning and multiplicity of sense have almost opposing semantic effects: while a homonym needs to be interpreted correctly in order to arrive at the correct meaning of an utterance, polysemes refer to different aspects or facets of the same concept, and might not even need to be completely specified to elicit a good-enough interpretation of what is meant \citep{Klepousniotou2002TheLexicon, pylkkanen2006representation, RECASENS20111138, Frisson_2015,poesio_forthcoming}. 
Evidence from psycholinguistic studies supports this distinction, indicating 
that polysemes are processed very differently than homonyms \citep{Frazier1990TakingSenses, Rodd_2002, Klepousniotou2008, Klepousniotou2012NotPolysemy}. 
A growing body of work recently also has started to challenge the uniform treatment of polysemic sense postulated by traditional theories such as the Generative Lexicon \citep{Pustejovsky1991, Asher2006, asher_2011}, and put forward proposals of a more structured mental representation of polysemic sense \citep{ortega2019}. Using 
co-predication tests, studies such as \citet{Antunes2003OnCo-Predication, traxler2005context, 10.3389/fpsyg.2013.00677} show that not all polysemic interpretations can be co-predicated, and that some sense interpretations  lead to zeugma:\footnote{Example from \citet{Cruse2009PolysemyViewpoint}} 
\begin{exe}
\ex
\# They took the door off its hinges and walked through it. 
\label{ex:polyseme_co-predication}
\end{exe}
Joining a range of recent work seeking to provide empirical data of graded word use similarity, in \citet{haber-poesio-2020-assessing, Haber2020} we recently released an experimental small-scale dataset of graded word sense similarity judgements for a highly controlled set of polysemic targets to investigate the notion of structured sense representations. Analysing results for ten seminal English polysemes, we observed significant differences among polysemic sense interpretations \cite{erk-etal-2013-measuring,nair-etal-2020-contextualized,trott-bergen-2021-raw} and found first evidence of a distance-based grouping of word senses in some of the targets. 

In this paper we present  
a modified and extended version of this initial dataset 
to 
i) provide additional annotated data to validate previous observations, and
ii) include new targets allowing for the same alternations as the initial set. This expansion enables us to carry out analyses not possible with the original dataset, including  iii) investigating similarity patterns and polysemy types,
iii) performing a detailed analysis the correlation between human judgements and sense similarities predicted by contextualised language models
, and 
iv) 
obtaining preliminary 
insights on how well their `off-the-shelf' representations of word sense can be used to cluster different sense interpretations of polysemic targets. 

The new data confirms previous observations of varying distances between polysemic word sense interpretations, and provides tentative evidence for similarity patterns within targets of the same type of polysemy. These patterns can to some degree be replicated by the similarities of embeddings extracted from BERT Large, opening up potential avenues of research utilising contextualised embeddings to proxy costly human annotations for the collection of a large-scale repository of fine-grained word sense similarity.  
The collected data is publicly available online.\footnote{\url{https://github.com/dali-ambiguity/Patterns-of-Lexical-Ambiguity}}

\section{Methods}
\label{sec:method}

The data for this study was created by revising and extending the dataset of contextualised word sense similarity presented in \citet{haber-poesio-2020-assessing,Haber2020}, and contains annotated sample contexts for different sense interpretations of 28 English polysemic nouns. 


\subsection{Target words}

Our initial dataset contained one target word for twelve frequently discussed types of logical metonymy \citep{doelling2018}. We focused on this form of \textit{regular} \citep{apresjan_1974, moldovan2019descriptions} or \textit{inherent} \citep{pustejovsky_2008} polysemy as it allows us to investigate and analyse the same interpretation patterns across a number of target word forms. We included the original data for eight targets, excluded two proper noun samples because their vanilla embeddings pooling sub-token encodings did not yield stable results under a simple cosine comparison, and re-collected annotations for two others that exhibited a high degree of annotation noise in the first collection effort. We then selected 18 additional targets for our second annotation effort, each allowing for the same alternations as one of the initial ten in order to investigate potential patterns in their distribution of sense interpretations. The new dataset contains the following set of seminal and experimental English polysemic target words: 

    \textbf{animal/meat}: lamb, chicken, pheasant, seagull;
    \textbf{food/event}: lunch, dinner;
    \textbf{container-for-content}: glass, bottle, cup;
    \textbf{content-for-container}: beer, wine, milk, juice;
    \textbf{opening/physical}: window, door;
    \textbf{process/result}: building, construction, settlement;
    \textbf{physical/information}: book, record;
    \textbf{physical/information/organisation}: newspaper, magazine;
    \textbf{physical/information/medium}:  CD, DVD;
    \textbf{building/pupils/directorate/institution}: school, university


\subsection{Sample sentences}

Following the approach detailed in \citet{Haber2020}, instead of collecting corpus samples containing the selected target words, custom samples were created such that i) the ambiguous target expression is the subject of the sentence, ii) the context is kept as short as possible, and iii) the context invokes a certain sense as clearly as possible without mentioning that sense explicitly.\footnote{As in ``The school is an old building." for sense \textit{building}. See \citet{Haber2020} for more details.}
With this method, pairs of sample sentences can easily be tested for target word similarity, as well as combined into co-predication structures to obtain acceptability judgements. 
As an example, polyseme \textit{newspaper} is traditionally assumed to allow for at least three sense interpretations: (1) \textit{organisation}, (2) \textit{physical object} and (3) \textit{information content}. In the materials, each of these  senses is invoked in two different contexts \textit{a} and \textit{b}: 
\vspace{0.5em}
\begin{enumerate}[nolistsep]
    \item [1a] The newspaper fired its editor in chief.
    \item [1b] The newspaper was sued for defamation.
    \item [2a] The newspaper lies on the kitchen table.
    \item [2b] The newspaper got wet from the rain.
    \item [3a] The newspaper wasn't very interesting.
    \item [3b] The newspaper is rather satirical today.
\end{enumerate}

\vspace{0.5em}

\noindent Comparing targets with the same number identifier results in what traditionally would be considered a same-sense scenario, and comparing targets with different number identifiers results in a cross-sense comparison.  
For co-predication, two contexts are combined into a single sentence 
by conjunction reduction \citep{zwicky1975ambiguity}. As an example, contexts 1a and 1b are combined into co-predication sample 1ab as follows: 
\vspace{0.5em}
\begin{enumerate}[nolistsep]
\item [1ab] The newspaper fired its editor in chief and was sued for defamation.
\end{enumerate}
\vspace{0.5em}

\noindent Besides polysemic alternations, some of the targets also allow for homonymic alternations (e.g. \textit{magazine} with different senses related to the print medium, but also a homonymic interpretation as a storage type). 
Feedback on homonymic interpretations will allow us to better put into perspective the results obtained for polysemic alternations. 

We omitted a collection of additional word class judgements  trialled in \citet{haber-poesio-2020-assessing} as we found that these judgements performed poorly in distinguishing polysemes from homonyms, and did not seem to exhibit the degree of sensitivity required for our analysis.   


    
    
\subsection{Human Annotation}

We collected human annotations online through Amazon Mechanical Turk (AMT). As a first measure of word sense similarity, we asked participants to rate the similarity in meaning of a target word shown in two different contexts --providing a meta-linguistic signal. Like in the initial data collection run, we did so by highlighting target expressions in bold font and asking annotators to rate the highlighted expressions using a slider labelled with ``The highlighted words have a completely different meaning'' on the left hand side and ``The highlighted words have completely the same meaning'' on the right. The submitted slider positions were translated to a 100-point similarity score, providing us with a graded word sense similarity judgement \citep{erk-etal-2013-measuring, Lau2007MeasuringJudgements}.
As a second measure of word sense similarity, we asked participants to rate the acceptability of a co-predication structure combining two contexts with the same target. We again used a slider, this time labelled with ``The sentence is absolutely unacceptable'' on the left and ``The sentence is absolutely acceptable'' on the right. In the co-predication setting, the polysemic target was not highlighted, providing us with a more ecological similarity judgement.

Annotators were paid 0.70 USD for a completed survey with 20 items, for an average expected hourly rate of 7.00 USD. To improve judgement quality, we required annotators to be located in the US, and have completed at least 5000 previous surveys with an acceptance rate of at least 90\%. Annotators judged items without any prior training based on minimal guidelines only.\footnote{For  full instructions and a screenshot of the annotation interface see Appendix \ref{app:instructions}} 

\subsection{Contextualised Language Models}

Models of polysemy have previously been proposed in distributional semantics \cite[see  for example][]{boleda-et-al:CL12}, but for the most part, such models found limited application in computational linguistics. This changed with the emergence of a new generation of contextualised language models like ELMo \cite{Peters2018DeepRepresentations}, BERT \cite{devlin2018bert} and GPT-2 \citep{radford2019language}, which led to impressive improvement in a number of NLP applications. In order to assess word sense similarity encoded in contextualised embeddings, we extracted target word embeddings from the different disambiguating contexts and calculated their cosine similarity (1-cosine). For ELMo we used the pretrained model on TensorFlow Hub\footnote{\url{https://tfhub.dev/google/ELMo/3}} and extracted target word vectors from the LSTM's second layer hidden state, which has previously been shown to encode the most semantic information \citep[see e.g.][]{Ethayarajh_2019, Haber2020}. We used the pretrained BERT Base (12 layers, hidden state size of 768) and BERT Large (24 layers, hidden state size of 1024) from the Huggingface transformers package.\footnote{\url{https://huggingface.co/transformers/pretrained_models.html}} As suggested by \citet{loureiro-jorge-2019-language}, we experimented with both the last hidden state and the sum of the last four hidden states as contextualised representation of a target word.\footnote{We also tested a pretrained implementation of GPT-2, but  excluded this model from our analysis, as due to its more traditional left-to-right text processing, all of our samples introducing targets as "The \textit{target}....", led to identical embeddings in different contexts.}
Lastly, we established a 
baseline 
by 
averaging over the static Word2Vec  \citep{Mikolov2013EfficientSpace} encodings of all words in a sample context to create a naive contextualised embedding.

\section{Results}
\label{sec:results}

In our analyses 
we focused
on three different aspects. 
First, we 
computed 
graded similarity and acceptability ratings based on the collected annotations,
and investigated how these ratings relate to traditional distinctions of lexical ambiguity and recent proposals of a more structured representation of polysemic senses, especially considering the patterns of word sense similarities observed across different target words allowing for the same set of sense alternations. 
We then analysed how the different contextualised language models' target embeddings correlate with either of the human annotations, and to what degree they 
replicate 
the patterns of word sense similarity observed in the human annotations. 
Lastly, we 
analysed 
the contextualised embbedings themselves, 
for a preliminary assessment
of how well these `off-the-shelf' word sense encodings fare in clustering samples based on their sense interpretation.  

\subsection{Word Sense Similarity Ratings}

\begin{figure}[t]
    \centering
    \includegraphics[width=0.49\linewidth]{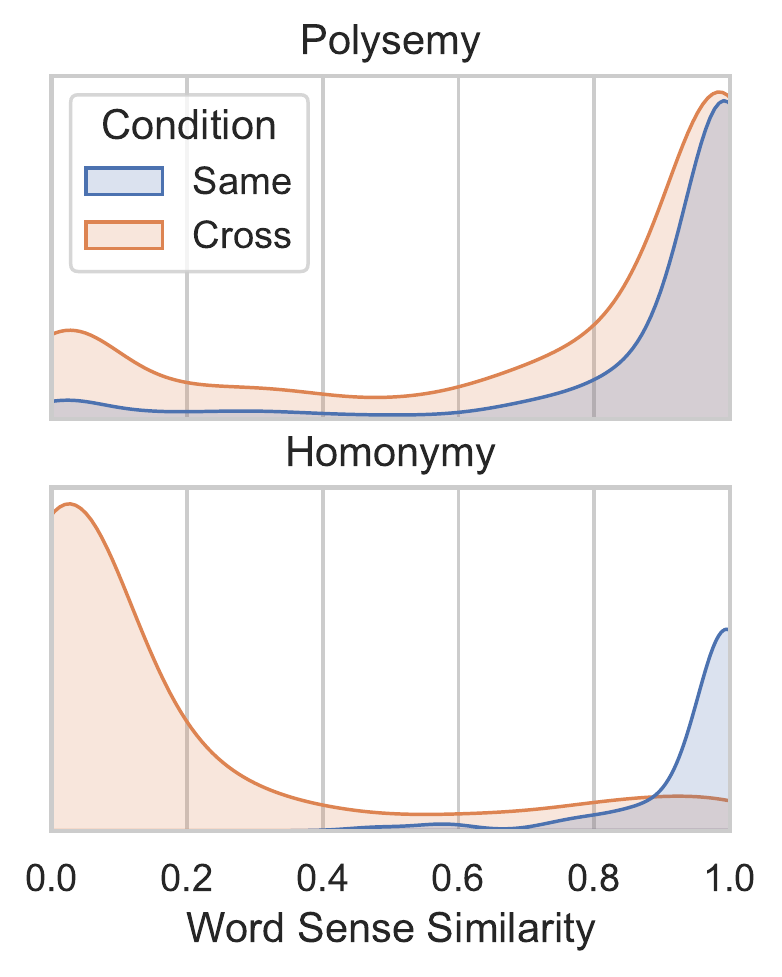}
    \includegraphics[width=0.49\linewidth]{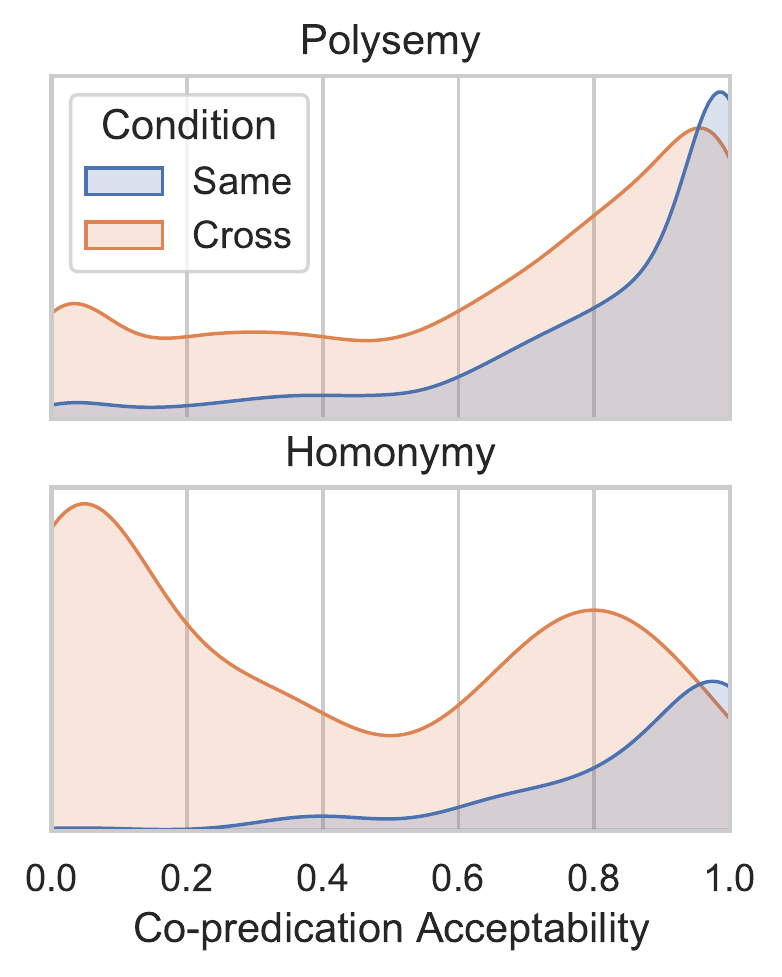}
    \caption{Distributions of explicit word sense similarity ratings and co-predication acceptability ratings given to same-sense (blue) and cross-sense (orange) samples with polysemic and homonymic alternations.}
    \label{fig:sim_distributions}
\end{figure}

\begin{figure*}[t]
    \centering
    \includegraphics[width=0.24\linewidth]{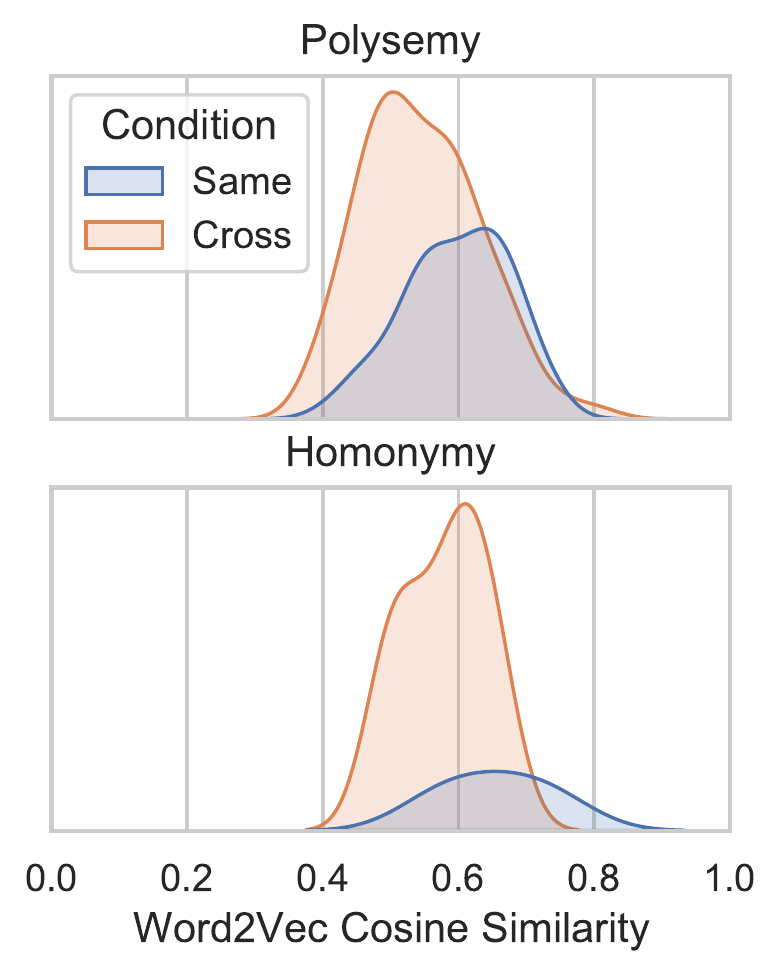}
    \includegraphics[width=0.24\linewidth]{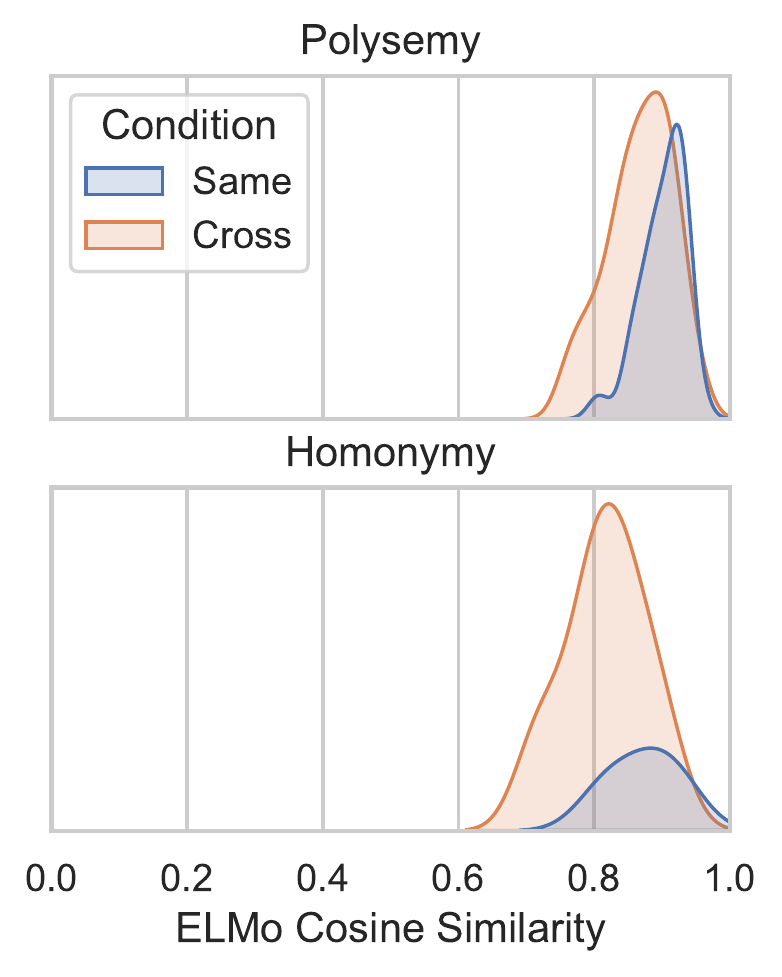}
    \includegraphics[width=0.24\linewidth]{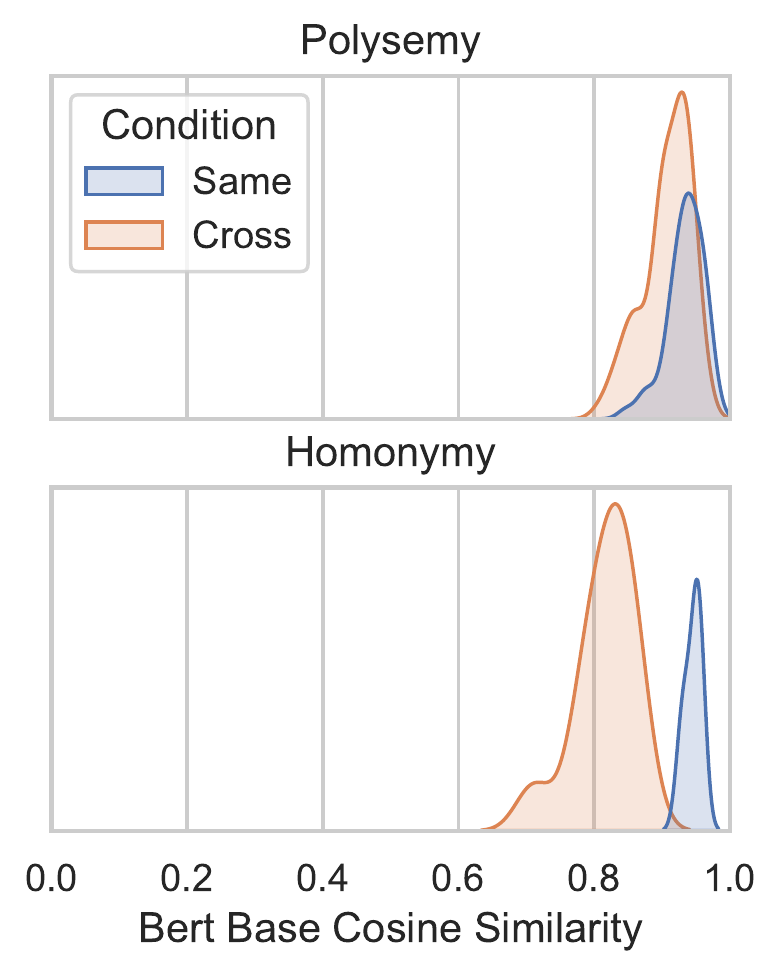}
    \includegraphics[width=0.24\linewidth]{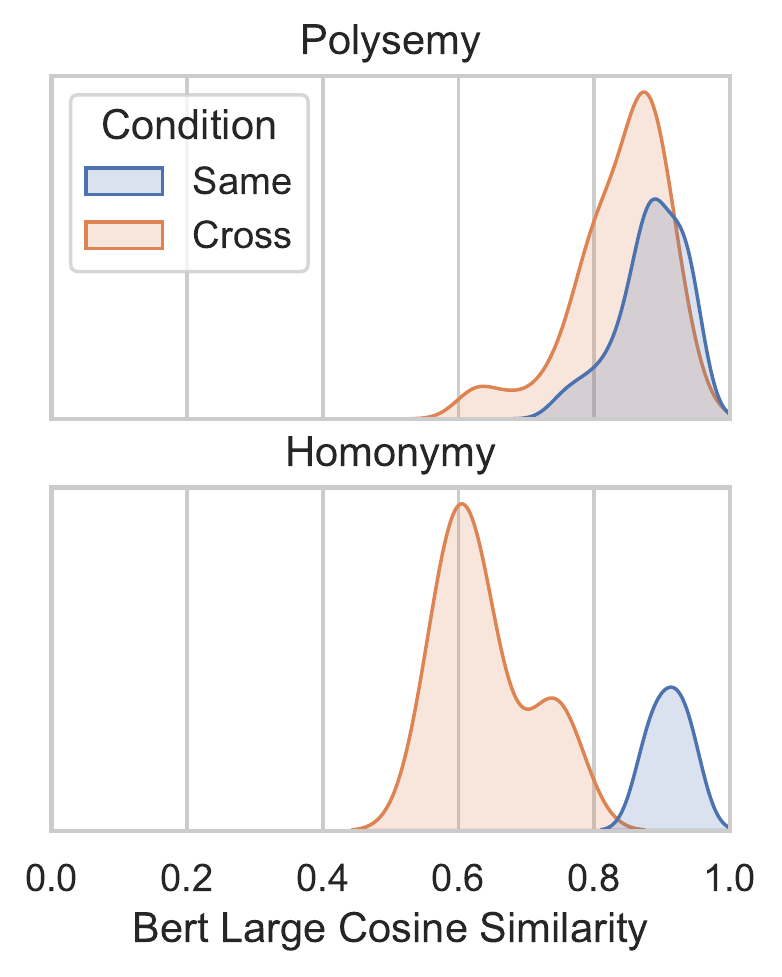}

    \caption{Distributions of embedding similarity scores obtained for same-sense (blue) and cross-sense (orange) samples with polysemic and homonymic alternations. BERT results for summing over the last four hidden states.}
    \label{fig:comp_dist}
\end{figure*}

In our second annotation effort, we collected an additional 8980 pairwise judgements
from 220 unique AMT participants rating the similarity of highlighted target words in different contexts. After filtering, we retained a total of 5862 judgements
\citep[including those of][]{Haber2020}, with an average of 16.5 annotations per item (minimum 7)\footnote{See Appendix \ref{app:filtering} for more details on filtering} and a per-questionnaire inter-annotator agreement rate of 0.62 \citep[Krippendorff's alpha,][]{arstein_2008} --which is relatively high considering the continuous rating scale provided to our annotators.

We first investigated potential effects of predicate ordering by applying a Mann-Whitney \textit{U} test \citep{mann-whitney} to ratings for identical context pairs that were presented in a different order during annotation. Only 22 of 229 pairwise tests yielded p-values <0.05, and none passed Bonferroni correction. We therefore concluded that --as expected-- predicate ordering effects are negligible for the explicit word sense similarity ratings based on our materials, and combined results for further analysis. 
Figure \ref{fig:sim_distributions} (left column) shows the distributions of word sense similarity ratings collected across all target words, separated on whether or not there is a sense alternation in the sample, and whether this alternation is traditionally considered to be polysemic or homonymic in nature. 
Homonymic cross-sense samples obtained a mean similarity rating of just 0.17, significantly lower than the overall same-sense mean of 0.89 (p-value <0.05). Polysemic cross-sense samples received a mean similarity score of 0.73, which is both significantly lower than the same-sense mean, 
and significantly higher than the homonym mean (see Table \ref{tab:comp_dist}, row 1). These results support 
the traditional view 
that polysemy occupies a distinctive middle ground between identity of meaning 
and homonymy 
\citep{Pinkal:1995}. 

Next, we grouped the data based on target words, and performed pairwise comparisons on all ratings given to their cross-sense interpretations. A large number of significant comparisons would indicate a high variance in the assigned ratings; a low percentage of significant differences indicates a consistent rating of samples. Due to the large number of tests, we then carry out a Bonferroni correction on the obtained results to establish a corrected, more conservative significance level and determine a lower bound on this statistic. Comparing all combinations of same-sense pairings for example, 20 of 58 tests yielded significantly different results (p-values <0.05), but only 4 entries passed Bonferroni correction (6.90\%), indicating that same-sense samples are quite consistently rated to invoke very similar interpretations. 
14.71\% of the 34 pairwise comparisons of homonymic cross-sense samples passed Bonferroni correction, as did 23.44\% of the 337 pairwise comparisons between ratings for polysemic cross-sense samples. Ratings for cross-sense samples therefore are less consistent than same-sense ratings, and polysemic alternations are rated more inconsistently than homonymic ones. Observing this variance in  similarity scores justifies our use of a continuous rating scale 
for the annotation experiments. With almost a quarter of the similarity ratings for polysemic sense alternations showing significant differences to those of other senses, these results also provide a novel type of empirical evidence against a uniform treatment of polysemic senses. 


\subsection{Co-Predication Acceptability Ratings}

\begin{table}[t]
    \resizebox{\linewidth}{!}{%
\begin{tabular}{|l|rrr|rrr|}
\hline
                    & \multicolumn{3}{c|}{\textbf{Same-Sense}}                                                                & \multicolumn{3}{c|}{\textbf{Cross-Sense}}                                                               \\
\textbf{Measure}    & \multicolumn{1}{l}{\textbf{Pol.}} & \multicolumn{1}{l}{\textbf{Hom.}} & \multicolumn{1}{l|}{\textbf{p}} & \multicolumn{1}{l}{\textbf{Pol.}} & \multicolumn{1}{l}{\textbf{Hom.}} & \multicolumn{1}{l|}{\textbf{p}} \\ \hline
Similarity          & 0.89                              & 0.96                              & 0.03                            & 0.73                              & 0.17                              & \textless{}0.05                 \\
Acceptability       & 0.83                              & 0.86                              & 0.10                            & 0.64                              & 0.41                              & \textless{}0.05                 \\ \hline
Word2Vec            & 0.60                              & 0.65                              & 0.12                            & 0.55                              & 0.58                              & 0.06                            \\
ELMo                & 0.90                              & 0.87                              & 0.14                            & 0.87                              & 0.82                              & \textless{}0.05                 \\
BERT Base           & 0.91                              & 0.93                              & 0.22                            & 0.88                              & 0.78                              & \textless{}0.05                 \\
BERT Base (L4)  & 0.93                              & 0.95                              & 0.27                            & 0.91                              & 0.82                              & \textless{}0.05                 \\
BERT Large          & 0.79                              & 0.85                              & 0.15                            & 0.72                              & 0.44                              & \textless{}0.05                 \\
BERT Large (L4) & 0.88                              & 0.91                              & 0.18                            & 0.84                              & 0.64                              & \textless{}0.05                 \\ \hline
\end{tabular}
}
\caption{Word sense similarity distribution means for the different measures investigated in this study. p-values calculated through Mann-Whitney \textit{U}.}
\label{tab:comp_dist}
\end{table}

\begin{table*}[t]
   \centering
    \resizebox{0.92\textwidth}{!}{%
\begin{tabular}{|ll|rr|rrrrrr|}
\hline
\multicolumn{2}{|l}{\textbf{Combination}}         & \multicolumn{2}{|l}{\textbf{Correlation}}                        & \multicolumn{6}{|l|}{\textbf{Ordinary Least Squares (OLS) Regression Analysis}}                                                                                                               \\
\hline
\textbf{First Measure} & \textbf{Second Measure} & \multicolumn{1}{l}{\textbf{r}} & \multicolumn{1}{l}{\textbf{p}} & \multicolumn{1}{|l}{\textbf{Coef.}} & \multicolumn{1}{l}{\textbf{R\textsuperscript{2}}} & \multicolumn{1}{l}{\textbf{F-stat.}} & \multicolumn{1}{l}{\textbf{Prob.}} & \multicolumn{1}{l}{\textbf{Omnib.}} & \multicolumn{1}{l|}{\textbf{Prob.}} \\
\hline
Similarity             & Acceptability           & 0.698                          & 1.09E-25                       & 0.484                              & 0.487                           & 156.571                              & 1.09E-25                           & 9.733                               & 0.008                                                   \\
Acceptability          & Similarity              & 0.698                          & 1.09E-25                       & 1.005                              & 0.487                           & 156.571                              & 1.09E-25                           & 0.967                               & 0.617                                                \\
\hline

Word2Vec               & Similarity              & 0.206                          & 0.008                          & 0.675                              & 0.042                           & 7.309                                & 0.008                              & 31.562                              & 0                                                           \\
Word2Vec               & Acceptability           & 0.311                          & 4.39E-05                       & 0.707                              & 0.097                           & 17.625                               & 4.39E-05                           & 9.668                               & 0.008                                 \\

ELMo                   & Similarity              & 0.515                          & 1.11E-12                       & 2.863                              & 0.265                           & 59.475                               & 1.11E-12                           & 10.43                               & 0.005                                                       \\
ELMo                   & Acceptability           & 0.523                          & 4.39E-13                       & 2.018                              & 0.273                           & 61.973                               & 4.39E-13                           & 6.552                               & 0.038                                                         \\

BERT Base               & Similarity              & 0.641                          & 1.02E-20                       & 4.070                              & 0.411                           & 115.185                              & 1.02E-20                           & 3.496                               & 0.174                                                     \\
BERT Base               & Acceptability           & 0.560                          & 3.43E-15                       & 2.469                              & 0.314                           & 75.521                               & 3.43E-15                           & 2.07                                & 0.355                                                      \\
BERT Large              & Similarity              & 0.687                          & 1.22E-24                       & 2.181                              & 0.472                           & 147.361                              & 1.22E-24                           & 15.96                               & 0                                            \\
BERT Large             & Acceptability           & 0.550                          & 1.40E-14                       & 1.212                              & 0.302                           & 71.520                               & 1.40E-14                           & 5.324                               & 0.07                                                      \\
\hline
\end{tabular}
}
\caption{Correlations between measures of contextualised word sense similarity. The first set of columns displays pairwise correlation based on Pearson's \textit{r}, the second set shows the key statistics obtained from an OLS regression analysis. BERT results for summing over the last four hidden states. }
\label{tab:ols_annotations}
\end{table*}

Besides these explicit similarity ratings, we collected an additional 
8640 judgements from 192 participants rating the acceptability of co-predication structures created from our sample sentences. 
After adding judgements for selected targets from the initial data and filtering noisy annotations, we retained a total of 
7379 judgements, for an average of 16.75 annotations per target word (minimum 12). 
Co-predication acceptability is meant to provide a more ecological signal of word sense similarity than the explicit similarity ratings,
with participants 
less 
aware of the factors that influence the perceived acceptability of the evaluated sentence. Per-questionnaire inter-annotator agreement here only reached a Krippendorff's alpha rating of 0.34, indicating stronger individual differences in the participants' use of the continuous rating scale. 

Investigating order effects in our co-predication samples revealed that only 1 of 229 pairwise comparisons between the acceptability scores of co-predication structures with different predicate orderings passed the Bonferroni corrected significance level of 0.00021.
We therefore argue that our samples are free from any secondary acceptability factors based on predication order \cite{murphy_2020}, and therefore indeed primarily test for the acceptability of invoking different senses of the target words. Based on this observation, we again combine results before further analysis. Figure \ref{fig:sim_distributions} (right column) shows the distributions of collected co-predication acceptability ratings split by sample condition and ambiguity type. The average acceptability rating for co-predication structures invoking the same sense in both predications is 0.83, the mean acceptability for homonymic cross-sense samples is 0.41, and the mean acceptability for polysemic alternations is 0.64 --significantly lower than the same-sense mean but significantly higher than the homonym mean (see Table \ref{tab:comp_dist}, row 2). 
These results support previous observations of co-predication acceptability, too, being a non-binary signal but rather forming a continuum \citep{Lau2007MeasuringJudgements} and  provide an additional challenge to 
co-predication as a linguistic test to distinguish polysemy from homonymy. Same-sense and homonymic samples were rated quite consistently, with only 10.34\% and 5.88\% of pairwise comparisons passing Bonferroni correction, respectively. Polyseme samples again show some degree of inconsistency, 
with 21.66\% of comparisons among polysemic cross-sense samples passing the corrected significance threshold of 0.00015. These results duplicate the observations made above, and provide additional evidence for the non-uniformity in interpreting polysemic samples.   

\subsection{Computational Ratings}

\begin{figure}[t]
    \centering
    \includegraphics[width=0.49\linewidth]{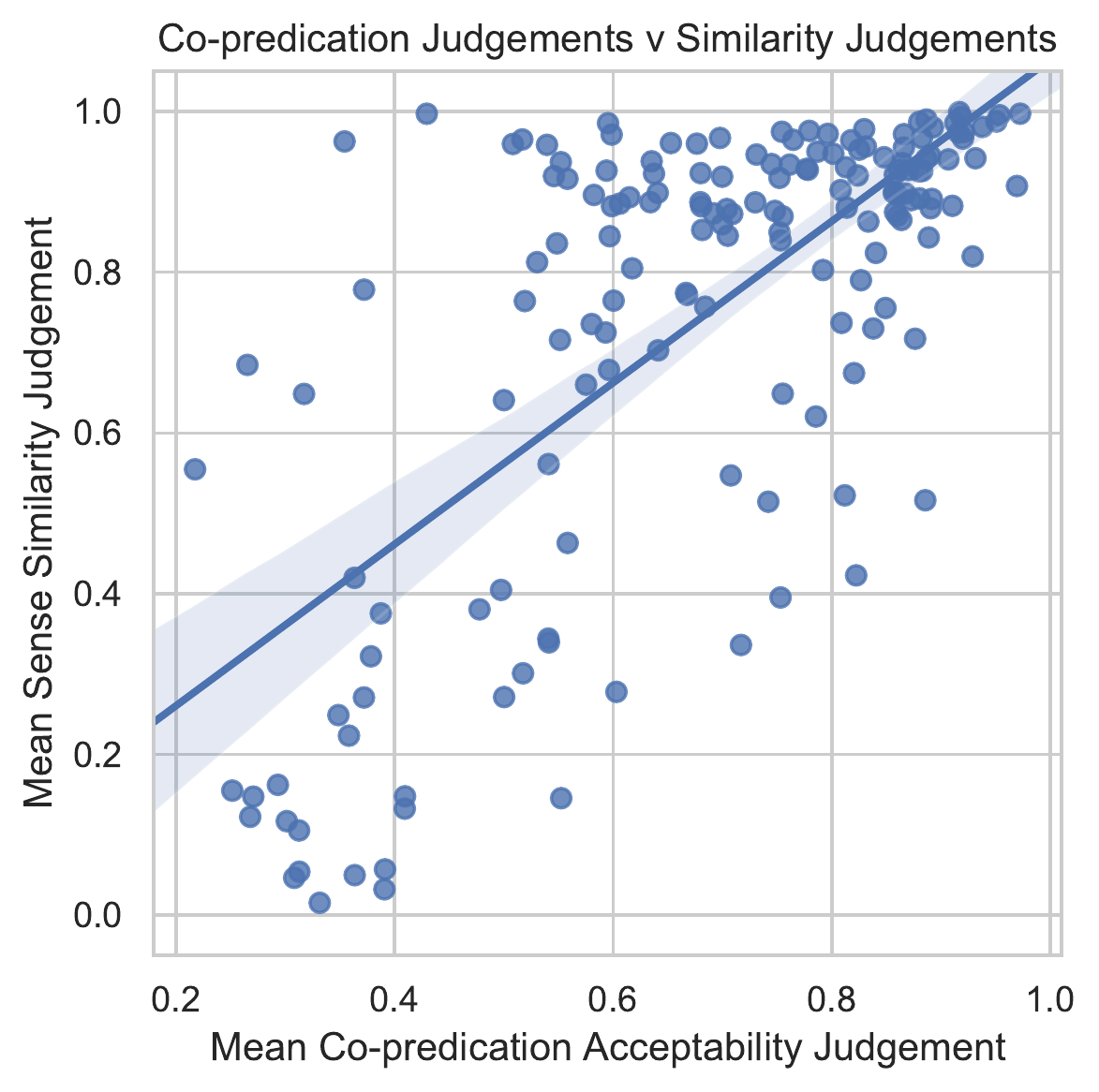}
    \includegraphics[width=0.49\linewidth]{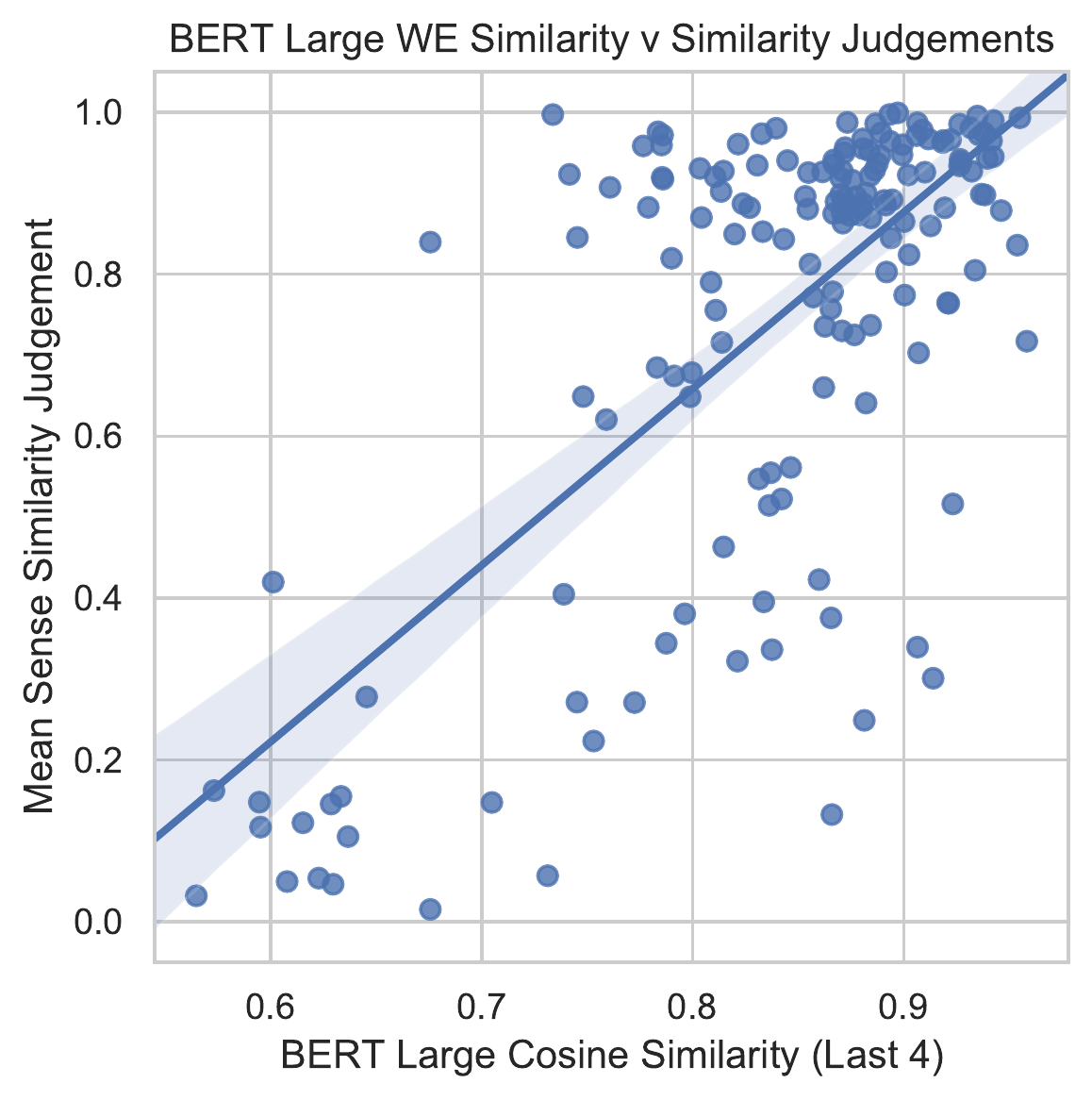}

    \caption{Correlations of co-predication v word sense similarity ratings (left) and BERT Large cosine similarity scores v  word sense similarity ratings (right), together with the best linear fit. 
    Scaling of x-axis adjusted for clarity. BERT results for summing over the last four hidden states.}
    \label{fig:correlations_annotations}
\end{figure}

We extracted contextualised embeddings of target word forms using the models described above, and determined pairwise similarity scores by calculating the embeddings' cosine similarity (1-cosine). As samples were encoded individually, there are no potential order effects here. Figure \ref{fig:comp_dist} visualises the distribution of target embedding similarity scores, and the bottom part of Table \ref{tab:comp_dist} details their distribution means. 
It is instantly noticeable that all computational
models
assign a much 
narrower 
range of similarity scores to the ambiguous samples --an observation 
already made 
in \citet{Ethayarajh_2019}. 
Homonymic and polysemic cross-sense ratings do not form significantly different distributions in the embeddings of the 
static Word2Vec model
(p-value 0.06), and --even more problematic-- homonymic cross-sense samples show no significant difference to same-sense samples (p-value 0.09). ELMo surprisingly struggles with the same distinction (p-value 0.09), but all BERT models produce clearly distinct distributions for polysemic, homonymic and same-sense samples (all p-values <0.05). 

In order to establish a measure of correlation between the similarity scores predicted by the contextualised models and the collected human judgements, we  calculated their pairwise correlation (Pearson's \textit{r}), and performed an ordinary least squares (OLS) regression for each combination of contextualised language model and human sense similarity measure. The results of these calculations are displayed in Table \ref{tab:ols_annotations}, and a selection of the pairwise comparisons is visualised in Figure \ref{fig:correlations_annotations}. 
The 
non-contextualised 
Word2Vec baseline displays a low but significant correlation with both human similarity measures, and shows an overall low goodness-of-fit, with R\textsuperscript{2} values of the OLS regression at 4\% and 10\%, respectively. 
ELMo clearly outperforms this baseline, both in terms of correlation with the human measures, as well as in its goodness-of-fit in the OLS regression analysis. For both BERT models, summing over the last 4 hidden states 
improved correlation with the similarity ratings by about 6 points, and the correlation with the co-predication acceptability ratings by about 4 points. We will therefore report only results on this version in the remainder of this paper.
Both models show a similar performance in predicting co-predication acceptability ratings as ELMo, with a slight lead by BERT Base, but BERT Large is clearly the best-performing model when predicting explicit similarity scores, with a correlation of 0.69 to the human annotation, and an R\textsuperscript{2} goodness-of-fit of 47\%. This high degree of correlation is also visible in the scatter plot in Figure \ref{fig:correlations_annotations}.
These results suggest that particularly BERT Large seems to be able to capture nuanced word sense distinctions 
in a similar way as
human annotators, and are in contrast to our initial findings reported in 
\citet{Haber2020}, where we measured 
a correlation of only 0.21 between BERT Base target embedding cosine similarities and word sense similarity judgements. We suggest that this difference in correlation is due to a number of factors, including i) the omission of the unstable proper noun targets, ii) the re-collection of annotations for particularly noisy items, iii) the use of a significantly larger amount of data, and iv) the inclusion of a number of homonymic targets, which populate the lower end of the spectrum and facilitate a better fit. 

\subsection{Similarity Patterns}
\label{sec:patterns}

\begin{figure*}[t]
    \centering
    \includegraphics[width=0.88\linewidth]{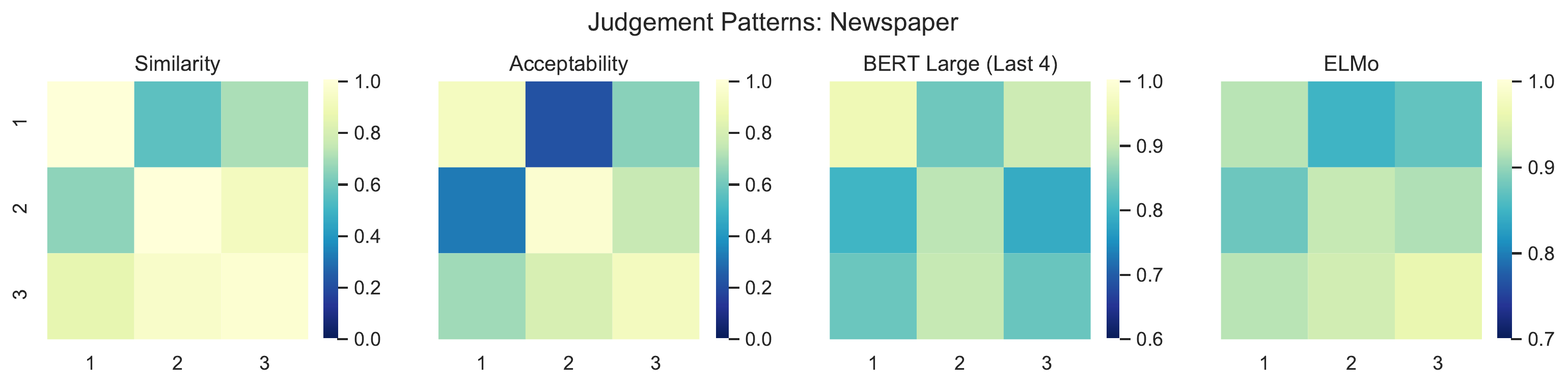}
    \includegraphics[width=0.88\linewidth]{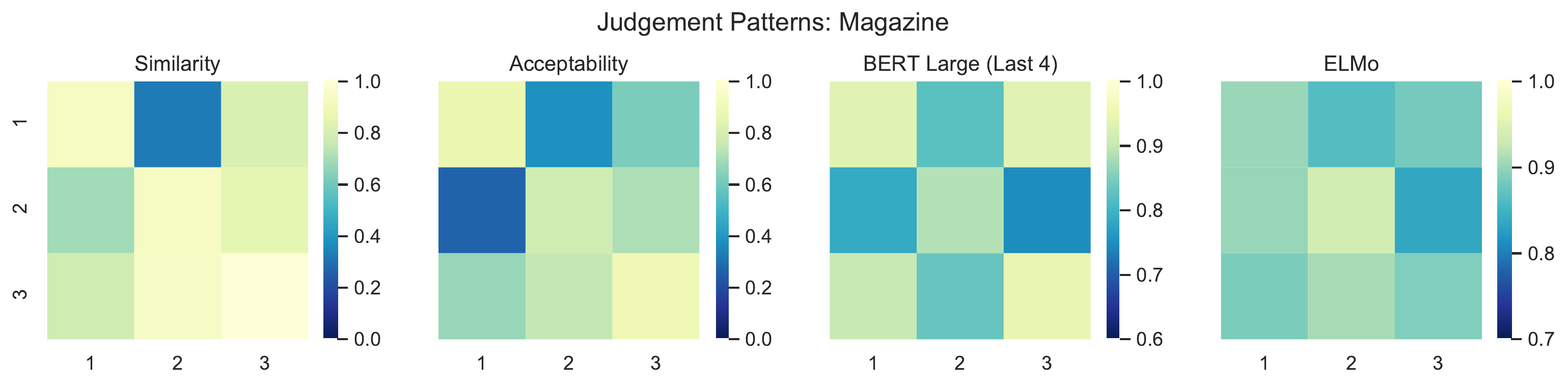}
    \caption{Similarity patterns in the sense similarity ratings for polysemes \textit{newspaper} and \textit{magazine}. \\ Senses: 1-physical, 2-information, 3-organisation. Colour scales adjusted for computational measures.}
    \label{fig:similarity_maps_newspaper_magazine}
\end{figure*}

\begin{table}[t]
    \centering
        \resizebox{\linewidth}{!}{%
 \begin{tabular}{|l|ll|ll|}
\hline
&  \multicolumn{2}{c}{\textbf{Pairwise}} & \multicolumn{2}{|c|}{\textbf{Overall}} \\
\textbf{Measure} & \textbf{\textit{r}} & \textbf{p \textless 0.05}  & \textbf{\textit{r}} & \textbf{p} \\ \hline
Similarity       & 0.44       & 3/24 (12.5\%)   &  0.53       &  8.260e-10\\
Acceptability    & 0.44     & 4/24 (16.7\%)      &   0.62    & 5.306e-14\\
ELMo             & 0.14     & 0/24 (0\%)           &   0.21 & 0.025\\ 
BERT Large       & 0.28       & 1/24 (4.2\%)      &  0.27    &  0.003 \\
\hline
\end{tabular}
}
    \caption{Mean Pearson correlation of polysemic word sense similarity patterns across different target words allowing the same alternation of senses, number of significant comparisons, and overall pattern correlation.}
    \label{tab:pattern_correlations}
\end{table}

\begin{table}[t]
    \centering
     \resizebox{\linewidth}{!}{%
    \begin{tabular}{|l|lll|lll|}
        \hline
        \textbf{Criterion}     & \textit{\textbf{t}} & \textbf{\#C} & \textbf{NMI} & \textbf{F1} & \textbf{P} & \textbf{R} \\ \hline
        Inconsistency & \textless{}0.7      & 3.54              & 0.60         & 0.77        & 0.86               & 0.71            \\
        Distance      & 31                  & 4.21              & 0.75         & 0.75        & 0.90               & 0.64            \\ \hline
    \end{tabular}
    }
    \caption{Best-performing settings for inconsistency and distance-based hierarchical Ward clustering of target word senses. \#C is the average number of clusters produced per target.}
    \label{tab:clustering_performance}
\end{table}

\begin{figure}[t]
    \centering
    \includegraphics[width=0.49\linewidth]{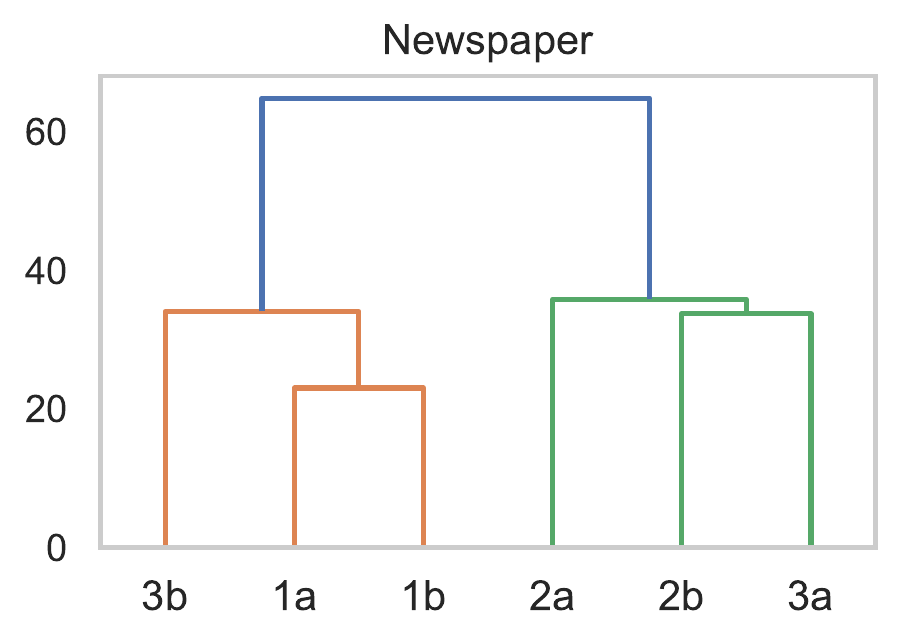}
    \includegraphics[width=0.49\linewidth]{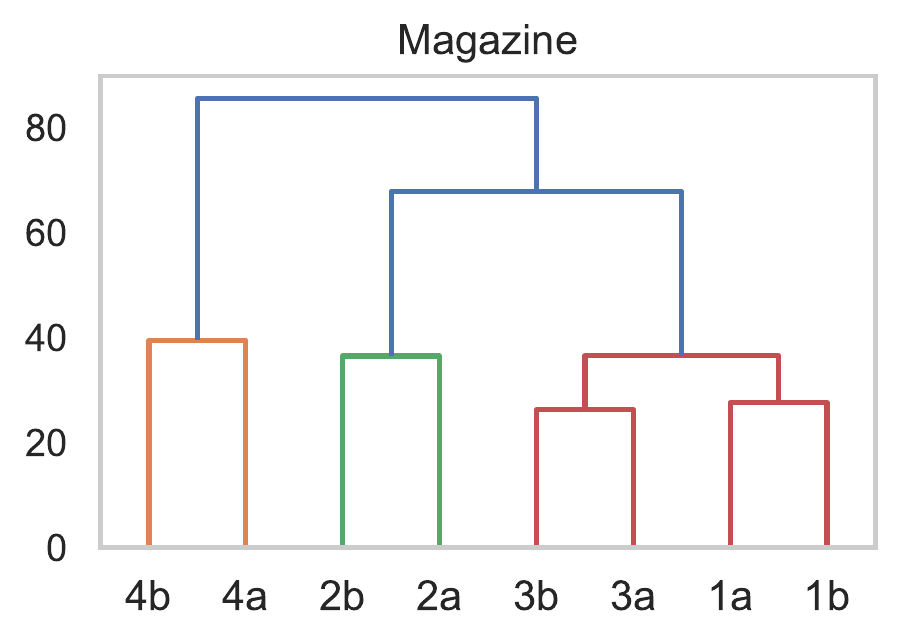}
    \caption{Dendograms of BERT Large contextualised embedding similarity for a selection of target words.
    Numbers indicate traditional sense distinctions.}
    \label{fig:dendograms_selection}
\end{figure}

One of the key 
reasons for 
extending 
our initial dataset 
was to add more target words for each of the tested types of polysemic sense alternation in order to allow for an investigation of sense similarity patterns across targets. 
Utilising the extended dataset, we established a set of 
similarity maps 
containing the mean similarity ratings for each combination of senses a given target word can take on, and compared these between targets of the same type. 
For example,
Figure \ref{fig:similarity_maps_newspaper_magazine} displays the similarity maps for target words \textit{newspaper} and \textit{magazine}. The correlation between these similarity maps reaches 0.89 (p-value = 0.001) in 
human 
similarity ratings, and 0.95 (p-value = 6.88e-05) for co-predication acceptability, indicating a clear pattern in the target's similarity ratings. In the similarity maps based on the cosines between BERT Large embeddings, the correlation reaches only 0.65 (p-value = 0.06), and just 0.34 (p-value = 0.37) in the ELMo similarity maps. 
The overall pattern correlations across target words of the same polysemy type can be found in
Table
\ref{tab:pattern_correlations}. 
The first set of scores are based on the correlations of all pairwise comparisons of polysemes that allow for the same alternations. Due to the small number of senses tested in this study, in most cases this comparison however does not allow for significant results. We therefore also calculated a second score by appending all pairwise comparisons into two separate lists and determining the correlation between these two lists. This is likely to represent a better estimate of overall pattern consistency, but might under-value inconsistent patterns. The mean correlation between BERT Large's similarity maps and the human sense similarity maps is 0.49, with one significantly similar pairing, and 0.52 compared to co-predication similarity maps (4 significant pairings) --rates comparable to the correlation between the two human annotations (mean \textit{r} = 0.54, 10 comparisons with p<0.05).

A qualitative analysis of the similarity maps revealed that while some alternation types like \textit{animal/meat} do exhibit consistent similarity patterns across targets, others like the \textit{content-for-container} alternation do not display any discernible similarity patterns at all.\footnote{See Figure \ref{fig:animal_meat} in the Appendix for the similarity maps of the \textit{animal/meat} targets} These observations suggest that sense similarity patterns are best to be investigated within a given type of alternation, and further research will be needed to develop a more detailed account of polysemy types and sense similarity patterns.     

 

\subsection{Sense Clustering}

As BERT Large displayed a high correlation with the human judgements of word sense similarity, and some capability in replicating similarity patterns across target words, we next wanted to investigate how well BERT's contextualised embeddings can be used to cluster our polysemous targets according to their interpretation \citep{mccarthy_2016, soler_2021}. To provide a tentative analysis, we grouped BERT Large's contextualised target encodings 
based on their similarity using the hierarchical Ward clustering method implemented in SciPy.\footnote{\url{ https://docs.scipy.org/doc/scipy/reference/generated/scipy.cluster.hierarchy.fcluster.html}} We opted for hierarchical clustering as this method has to determine the optimal number of clusters itself, and does not take this number as an argument like most clustering methods do. We experimented with two different clustering criteria based only on a threshold parameter \textit{t}.
The quantitatively best-performing settings are displayed in Table \ref{tab:clustering_performance}.\footnote{See Appendix \ref{app:clustering} for more detail on clustering} Both settings produce more clusters than the traditional grouping of the tested targets would assume, which indicates that especially precision scores might be artificially high --but overall the clustering seems to produce sensible results. Figure \ref{fig:dendograms_selection} displays a selection of dendograms produced by the clustering. 
The grouping of \textit{newspaper} interpretations clearly separates the \textit{organisation} sense 1 from the \textit{physical object} interpretation 3, but splits the \textit{information} samples 2 among the two, indicating the similarity in their contextualised embeddings. For magazine, the clustering of samples creates four clear groupings, with the \textit{organisation} reading showing the most similarity with the \textit{information} interpretations, and clearly separating the three polysemic senses from the homonymic \textit{storage} reading 4. The clustering of alternations like \textit{food/event}, \textit{animal/meat} and \textit{process/result} appears work consistently well, while others like the \textit{content-for-container} alternation lead to consistently wrong sense groupings.

\section{Related Work}


Most work focused on the word sense disambiguation capabilities of contextualised language models investigates the classification of homonyms with clear-cut evaluation criteria (see \citealt{wsd_2021} for a recent summary). Polysemy proper adds another dimension of difficulty, since related senses can be perceived to be very similar to one another, and some form of graded relatedness criterion in necessary to properly evaluate model predictions \citep{erk-etal-2013-measuring, Lau2007MeasuringJudgements}. Datasets that capture graded similarity judgements usually do so for word pairs in isolation --often intended to evaluate static word sense embeddings \citep{taieb2019survey}, or are conducted on a small number of items \citep{erk-etal-2013-measuring}. Notable exceptions are the Word in Context (Wic) dataset by \citet{pilehvar-camacho-collados-2019-wic}, which contains over 7,000 sentence pairs with an overlapping English word, but was annotated based on a binary classification task. The CoSimLex dataset \citep{armendariz2019cosimlex} on the other hand collects graded similarity judgements 
--but does so for different, related targets. 

In parallel to our work, \citet{nair-etal-2020-contextualized} recently conducted an investigation of 32 polysemic and homonymic word types extracted from the Semcor corpus \citep{miller_semcor} by comparing the distances between a selection of cross-sense samples as determined by participants arranging them in a 2D spatial arrangement task. 
In line with our results, they reported polysemic senses to be rated significantly more similar to one another in both the human annotations and BERT Base embeddings, and found a strong correlation between the cosine distance of BERT sense centroids and aggregated relatedness judgements. 
In a similar approach, \citet{trott-bergen-2021-raw} recently presented a novel dataset of 112 polysemes and homonyms, for each of which a number of highly controlled sentence pairs were annotated for similarity of use. While their data is very similar to ours, one noticeable difference can be found in the distribution of cross-sense polyseme ratings. Based on our samples, different polyseme interpretations were rated to be mostly quite similar still, but their data displays an almost even distribution of similarity scores assigned to them. A closer inspection of the targets used in their study revealed two main factors that are likely to have contributed to this difference. Firstly, while all of our targets were specifically chosen to be regular, metonymic polysemes, a large part of \citeauthor{trott-bergen-2021-raw}'s polysemic targets are examples of metaphoric polysemy. Re-analysing their data after distinguishing these different branches of polysemy might help to further investigate their respective effects. And secondly, we noticed the use of compound nouns (i.e. ice \textit{cone} vs traffic \textit{cone}) to disambiguate target words. Considering polysemy as a form of under-specified language use, we argue that these expressions might undermine the raison d'être of polysemy proper as they over-specify the ambiguous target --but highlight an interesting additional facet of this research.   



\section{Conclusion}

We present a revised and extended dataset of graded word sense similarity 
for 28 seminal, lexically ambiguous word forms. 
The collected data supports previous observations of significant similarity differences between 
polysemic interpretations 
and 
led to the discovery of tentative patterns of word sense similarity for certain types of alternations. While more work on this matter will be needed before definite conclusions can be drawn, both of these observations can be taken as additional 
evidence against linguistic models proposing a uniform treatment of polysemic senses. 
We also used the collected data to test how well different `off-the-shelf' contextualised language models can predict 
human word sense similarity ratings. Among the 
tested models, 
especially BERT Large seems to capture nuanced word sense distinctions in a similar way to human annotators, and to some degree is capable of grouping sense interpretations by their contextualised embeddings. 
We hope to further expand the dataset presented in this paper to create a novel, more complex benchmark for the word sense disambiguation (WSD) task. In this endeavour, contextualised language models could be used to automatically detect relevant target word forms, and to collect corpus samples exhibiting specific targets and interpretations to be rated by human annotators for a more realistic, real-world test bed.  




\section*{Acknowledgements}

The work presented in this paper was supported by the DALI project, ERC Grant 695662. Janosch Haber is now part of the Turing Enrichment Scheme. The authors would like to thank Derya \c{C}okal and Andrea Bruera for their input, and the anonymous reviewers for their feedback. 

\bibliography{custom}
\bibliographystyle{acl_natbib}


\appendix

\section{Annotation Instructions and Interface}
\label{app:instructions}

In the word sense judgement task, participants were given the following set of instructions: \\

"Carefully read each pair of sentences and specify how similar the highlighted words are by using the slider. The slider ranges from 'The highlighted words have a completely different meaning' on the far left to 'The highlighted words have completely the same meaning' on the far right. \\

There are 20 sentence pairs. \\

The survey contains a number of test items that can be used to determine whether you are carefully reading the sentences or are submitting random answers. Submissions that fail the test items will be rejected." \\

\begin{figure*}
    \centering
    \includegraphics[width=0.7\linewidth]{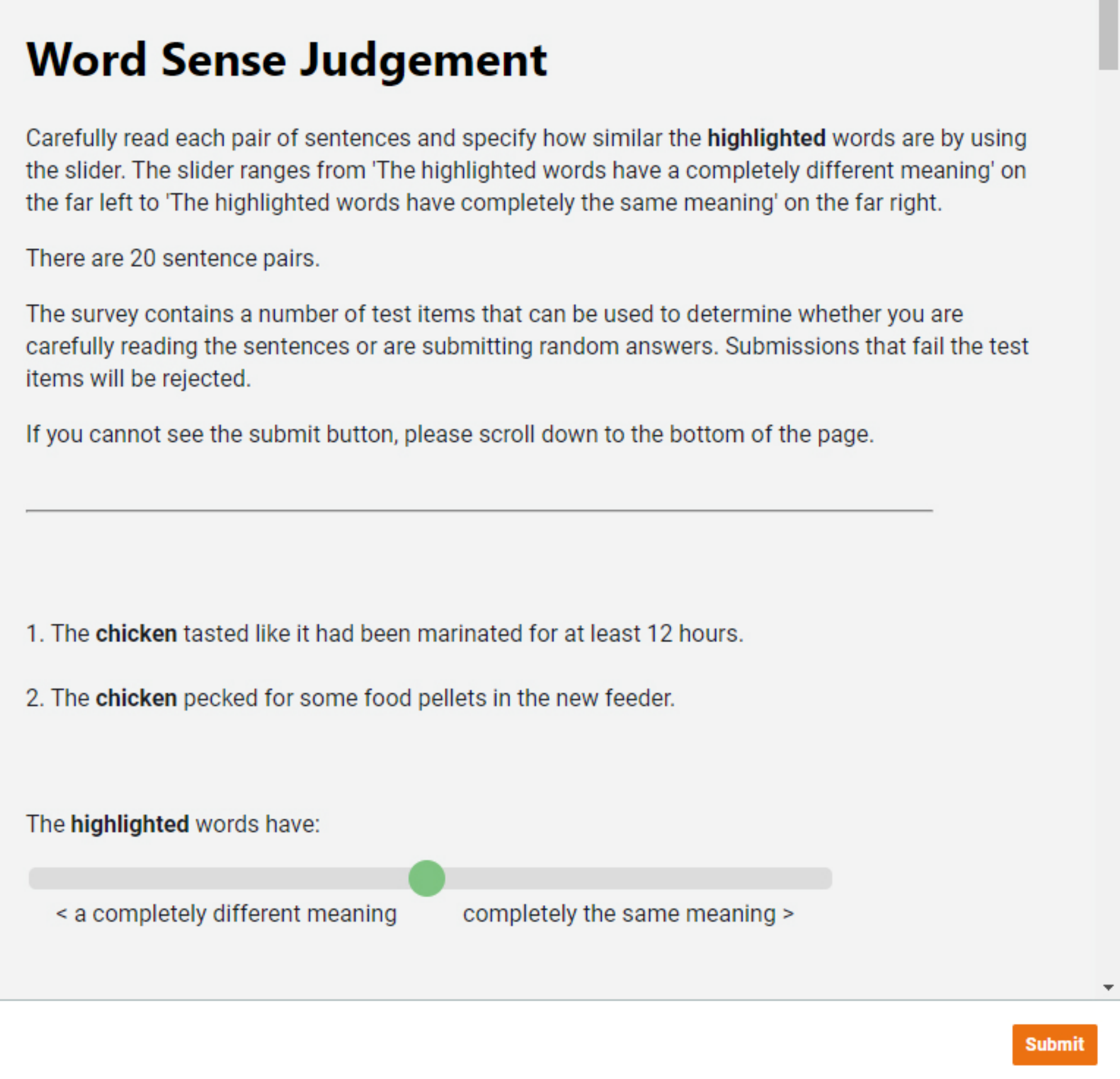}
    \caption{Screenshot of the the AMT interface for the explicit word sense similarity annotation task.}
    \label{fig:amt_screenshot}
\end{figure*}

A screenshot of the the AMT interface for this task is displayed in Figure \ref{fig:amt_screenshot}. In the sentence acceptability task, the following instructions were shown to the participants:\\

"Carefully read each sentence and specify how acceptable it is by using the slider. The slider ranges from 'The sentence is absolutely unacceptable' on the far left to 'The sentence is absolutely acceptable' on the far right.\\

There are 20 sentences.\\

The survey contains a number of test items that can be used to determine whether you are carefully reading the sentences or are submitting random answers. Submissions that fail the test items will be rejected."\\




\section{Filtering}
\label{app:filtering}

In order to reduce annotation noise, we filtered out submissions from participants who failed to rate test items according to a set of custom criteria. Surveys in both experiments each contained two test items.

In the word sense similarity annotation study, one test item would show a (homonymic) target interpreted in the same way in both sentences, with minimal changes to the context (\texttt{test-same}):

\begin{exe}
\ex
1. The \textbf{mole} dug tunnels all throughout the garden. \\
2. The \textbf{mole} dug tunnels under the flower bed.
\label{ex:test_s}
\end{exe}

The second test item would include two sentences with unrelated (homonymic) targets (\texttt{test-random}):

\begin{exe}
\ex
1. The \textbf{model} wore a new dress designed by Versace. \\
2. The \textbf{seal} indicated that the letter had never been opened. 
\label{ex:test_r}
\end{exe}

Submissions were excluded from analysis if either the \texttt{test-same} item was rated below 0.7 similarity, or the \texttt{test-random} item was rated above 0.2 similarity.

In the co-predication study, the \texttt{test-same} item would be no actual co-predication structure (to prevent any potential infelicitous co-predication), but a similar-looking sentence with a conjunctive phrase:

\begin{exe}
\ex
A group of boys were playing Frisbee in the park and a girl tried to balance on a slack line.
\label{ex:c_test_s}
\end{exe}

The \texttt{test-random} item would have the first part of a conjunctive sentence, but end it a randomly scrambled phrase:

\begin{exe}
\ex
The match ended without a clear winner and the off the managed bass hook get to.
\label{ex:c_test_r}
\end{exe}

Submissions were excluded from analysis if \textit{both}, the \texttt{test-same} item was rated below 0.7 similarity and the \texttt{test-random} item was rated above 0.2 similarity.

\section{Animal/Meat Similarity Maps}
\label{app:similarity_maps}

\begin{figure*}[t]
    \centering
    \includegraphics[width=0.98\linewidth]{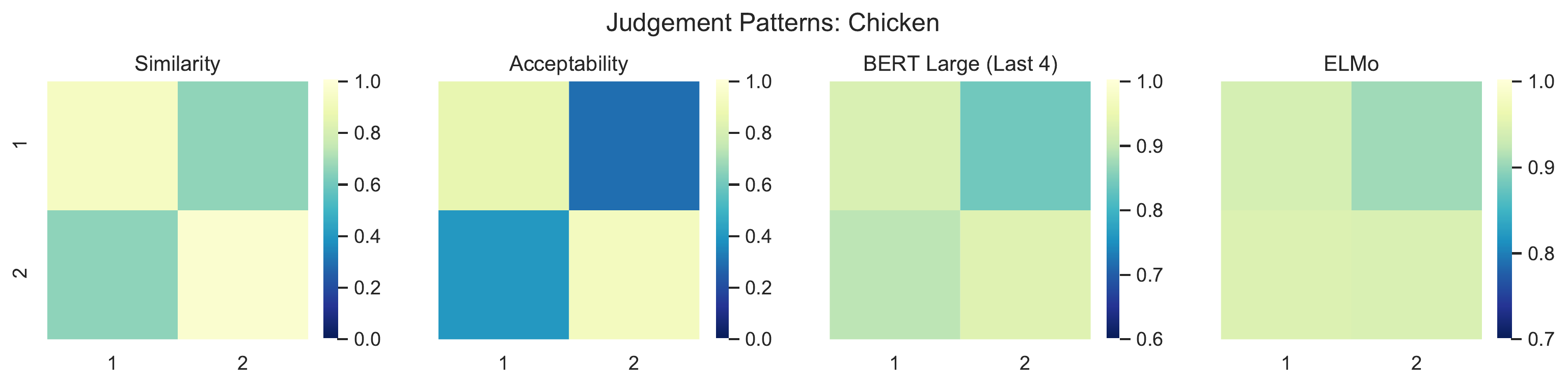}
    \includegraphics[width=0.98\linewidth]{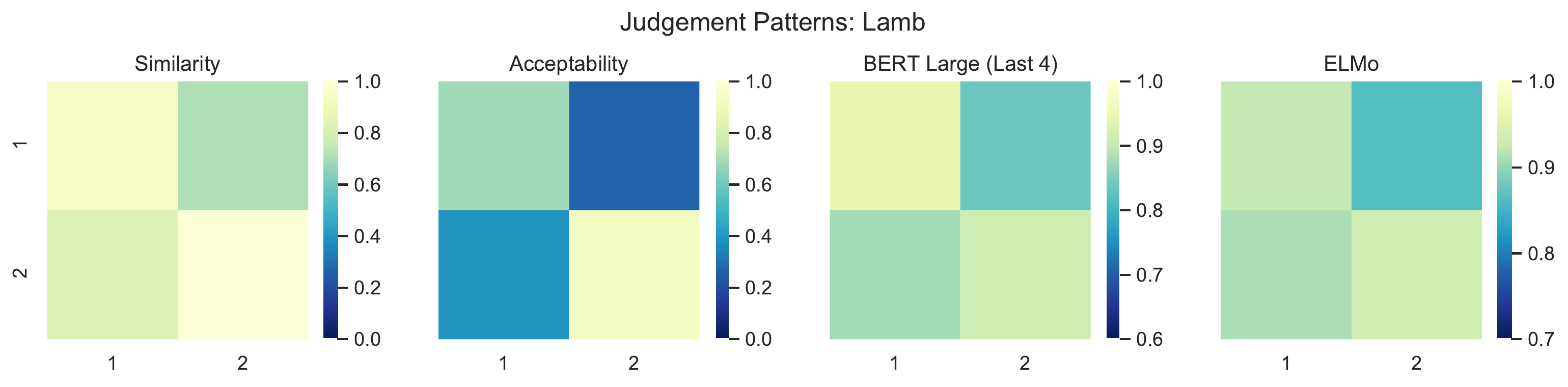}
    \includegraphics[width=0.98\linewidth]{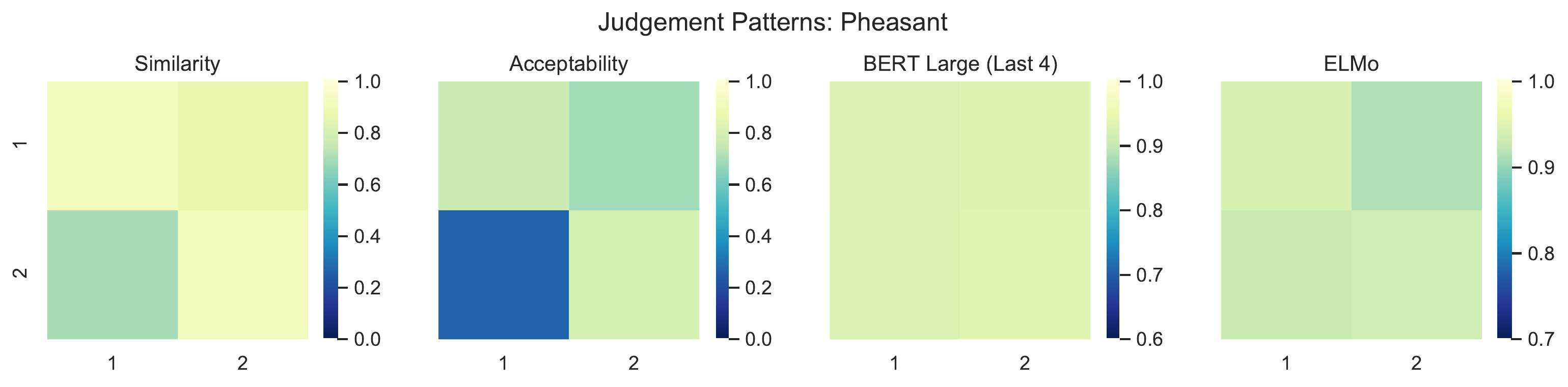}
    \includegraphics[width=0.98\linewidth]{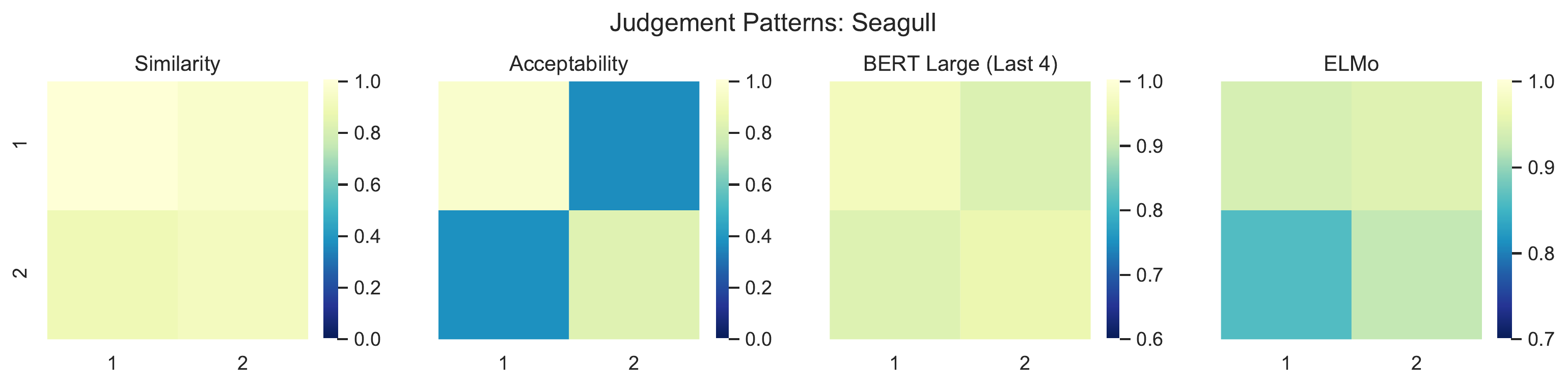}
    \caption{Similarity patterns in the sense similarity ratings for \textit{animal/meat} alternation polysemes. \\ Senses: 1-animal, 2-meat. Colour-scales adjusted for computational measures.}
    \label{fig:animal_meat}
\end{figure*}

Figure \ref{fig:animal_meat} shows the similarity maps for the tested animal/meat alternation polyseme targets \textit{chicken}, \textit{lamb}, \textit{pheasant} and \textit{seagull.} Both, \textit{chicken} and \textit{lamb} are common variants, \textit{pheasant} is less frequent, and \textit{seagull} would typically not be considered a member of this type, but still shows a similar pattern in the co-predication acceptability ratings and BERT Large cosine similarity. 

\section{Clustering}
\label{app:clustering}

\begin{figure*}[t]
    \centering
    \includegraphics[width=0.32\linewidth]{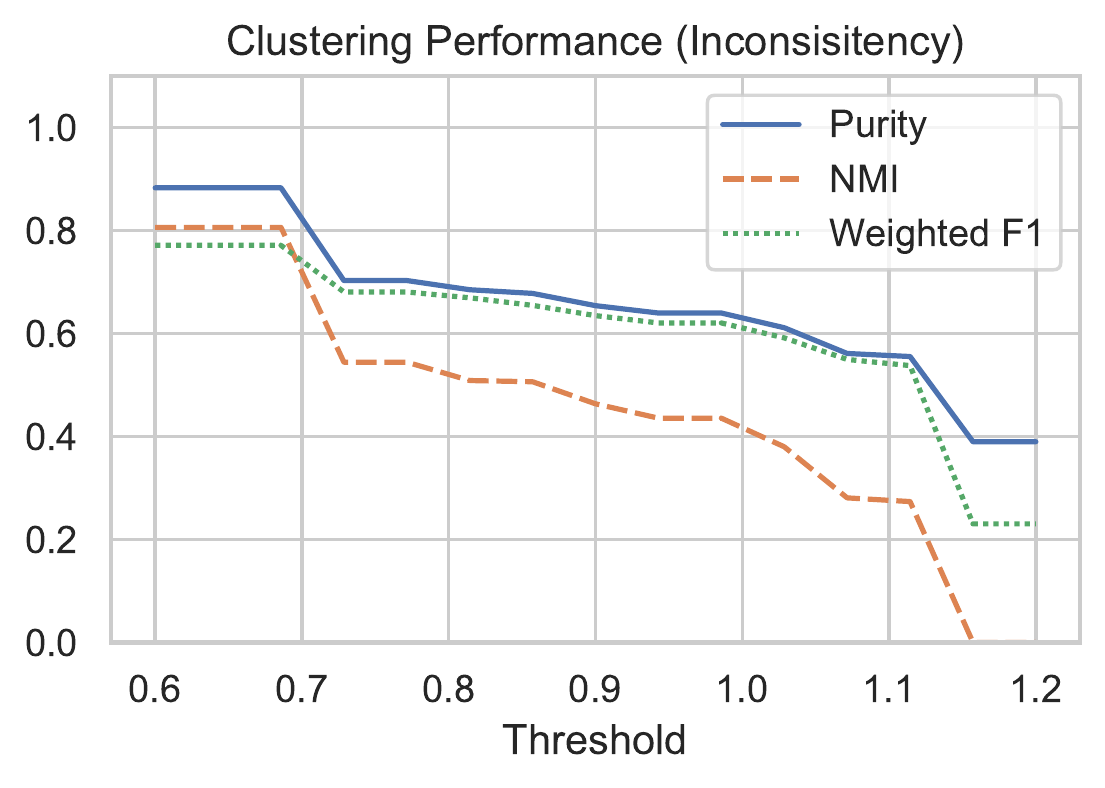}
    \includegraphics[width=0.32\linewidth]{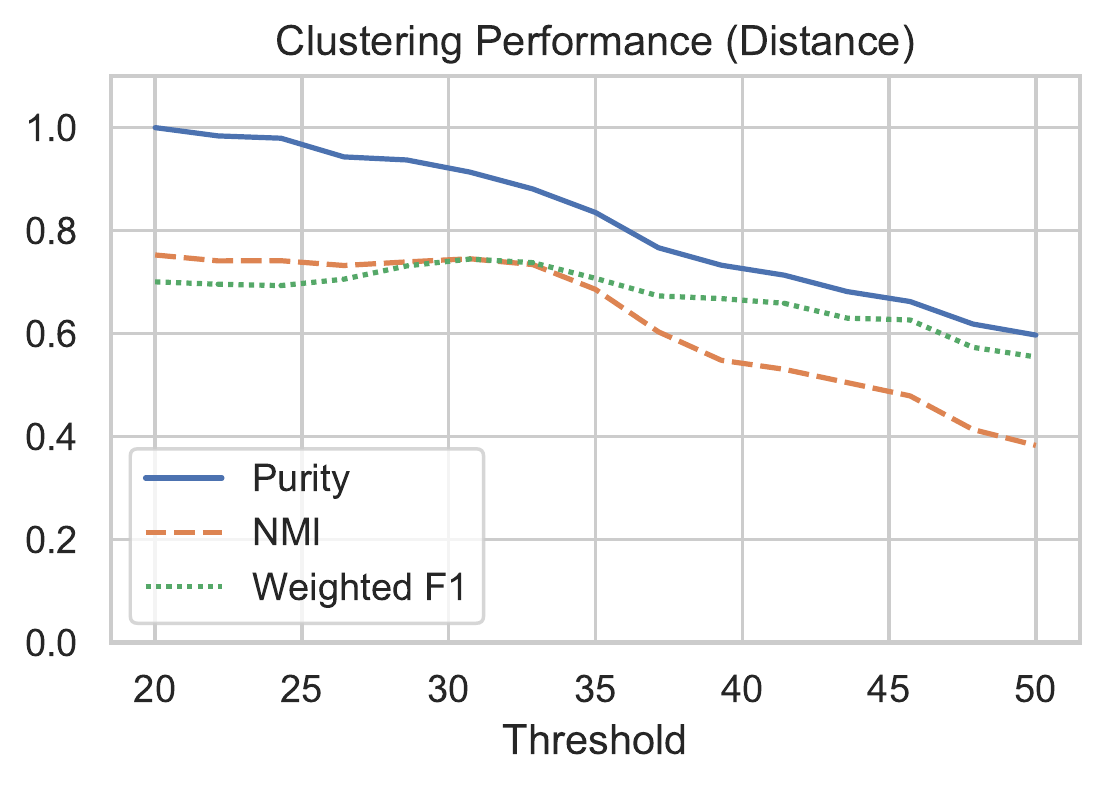}
    \includegraphics[width=0.34\linewidth]{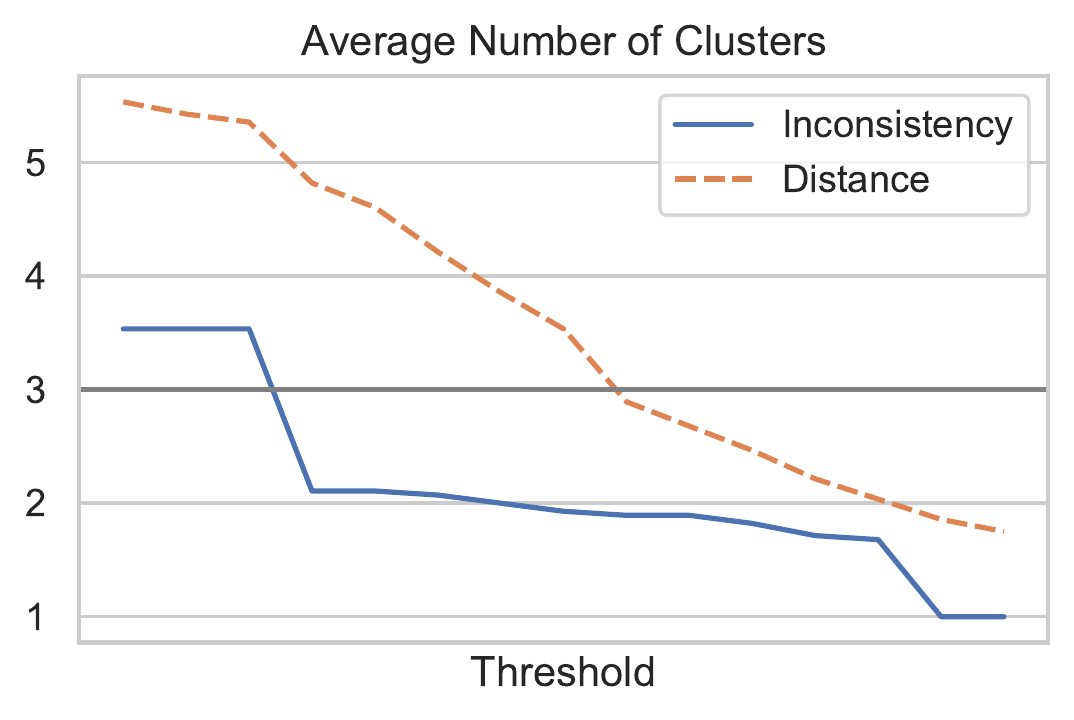}
    \caption{Clustering performance for the inconsistency (left) and distance (centre) criterion when grouping BERT Large contextualised embeddings with linear Ward clustering based on clustering threshold \textit{t}. Right: Average number of clusters produced by the clustering methods (gold mean: 3.0).}
    \label{fig:clustering_performance}
\end{figure*}

We experimented with two clustering criteria: using \textit{node inconsistency}, all leaf descendants of a cluster node belong to the same cluster if that node and all these descendants have an inconsistent value less than or equal to a threshold value \textit{t}. Under the \textit{distance} criterion, clusters are formed so that the observations in each cluster have no greater distance than the set threshold value \textit{t}. 
Figure \ref{fig:clustering_performance} shows the development of cluster purity, Normalised Mutual Information (NMI) and weighted F1 scores for different values of threshold \textit{t} using the inconsistency criterion (left) and distance criterion (centre). The right graph plots the average number of clusters produced by both measures with increasing threshold \textit{t} (gold mean: 3.0).

\end{document}